\documentclass[10 pt, conference]{ieeeconf}  % Comment this line out
\IEEEoverridecommandlockouts                              % This command is only
\usepackage{graphics} % for pdf, bitmapped graphics files
\usepackage{graphicx}
\usepackage{url,soul}
\usepackage[table]{xcolor}
\usepackage{array}

\usepackage[tracking=true]{microtype}

\title{\LARGE \bf
       Information Extraction  based on Named Entity for Tourism Corpus
}

\author{ \parbox{3 in}{\centering Chantana Chantrapornchai\\
        \thanks{978-1-7281-0719-6/19/\$31.00~\copyright 2019 IEEE.}    
        Dept. of Computer Engineering\\
         Faculty of Engineering\\
         Kasetsart University\\
         Bangkok, Thailand\\
         {\tt\small fengcnc@ku.ac.th}}
         \hspace*{ 0.5 in}
         \parbox{3 in}{ \centering  Aphisit Tunsakul\\
         Dept. of Computer Engineering\\
          Faculty of Engineering\\
         Kasetsart University\\
         Bangkok, Thailand\\
          {\tt\small apisit.n@ku.th}}
}

\begin{document}

\maketitle
\thispagestyle{empty}
\pagestyle{empty}

%%%%%%%%%%%%%%%%%%%%%%%%%%%%%%%%%%%%%%%%%%%%%%%%%%%%%%%%%%%%%%%%%%%%%%%%%%%%%%%%
\begin{abstract}

Tourism information is scattered around nowadays. To search for the information, it is usually time consuming to browse through the results from search engine,  select and view  the details of each accommodation.
 In this paper, we present  a methodology to extract  particular  information  from full text returned from the search engine to facilitate the users. Then, the users can specifically look to the desired relevant information.   The approach can be used for the same task in other domains.
 The main steps are 1) building training data and 2) building recognition model.
First, the tourism data  is gathered and the vocabularies are built. The raw corpus is used to train for creating  vocabulary embedding. Also, it is used for creating annotated data. The process of creating named entity annotation is presented.
Then, the recognition model of a given entity type can be built. 
From the experiments,  given hotel description,   the model can extract the desired entity,i.e, name, location, facility. The   extracted data  can further be stored as a structured information, e.g., in the ontology format, for future querying and inference. The model  for automatic named entity identification, based on machine learning, yields the error ranging 8\%-25\% . 

\end{abstract}

%%%%%%%%%%%%%%%%%%%%%%%%%%%%%%%%%%%%%%%%%%%%%%%%%%%%%%%%%%%%%%%%%%%%%%%%%%%%%%%%
\section{Introduction}
Typical information search in the web requires the text or string matching. When the user searches the information, the search engine returns the relevant documents that  contain  the matched string.  The users need to browse through the associated link to find whether the web site is in the scope of interest, which is very time consuming.

To facilitate the user search,
using ontology representation can enable the search to return  precise results. The specified keyword may refer to the meaning in the specific domain. For example,  consider the word, ``clouds".  The typical search matching such a keyword returns the documents referring to similar word such as ``sky". However, when using as ``cloud computing", the meaning is totally different. 
Also, with the capability of ontology, it can also infer to other relevant information. For example, "cloud computing" is a sub-field under ``computer architecture" . The relevant documents may include the paper in the area such as ``operating system", ``distributed system" etc.
 The proper ontology construction and imported data can lead to the enhanced search features.

It is known that for a given document, extraction data into the ontology usually required lots of  human work. Several previous works have attempted to propose methods for building ontology based on data extraction \cite{CC2016}.   Most of the work relied on the web structure documents \cite{Fei2009,Alani,Jak2007}. The ontology is extracted based on HTML web structure, and the corpus is based on
WordNet. For these approaches, the time consuming process is the  annotation which is to annotate the type of name  entity.
In  this paper, we target at the tourism domain, and  aim to   extract  particular information helping  for ontology data acquisition.

We present the framework for the given  named entity  extraction. Starting from the web information scraping process, the data are selected based on the HTML tag for corpus building. The data is used for model creation for automatic named entity recognition. The annotation for training data is also based on the tagged corpus.    The inputs of model is the sentences along with the tagged   entities. We also create  word embedding for our domain. The embedding represents the similarity degree of the vocabularies. The embedding can be used to other NLP tasks with these new words such as text summarization.

\section{Backgrounds}

In this section, we divide the backgrounds into subsections: first, we give examples of  the existing tourism ontology. Next, since we focus on the use of machine learning to extracting relations from documents, we describe the previous work in machine learning and deep learning in natural language processing. 
\subsection{Tourism ontology}
Lots of tourism ontology  were proposed. For example, Mouhim et al.  utilized the knowledge management approach for constructing ontology  \cite{Mou2011}.  
They created Morocco tourism ontology. 
The approach considered Mondeca tourism ontology in OnTour \cite{Sior2004} proposed by Siorpaes et al. They built the vocabulary  from thesaurus obtained from the United Nation World Tourism Organisation (UNWTO). The  classes as well as social platform were defined.
In \cite{sigala2007}, the approach for building e-tourism ontology is  :
 NLP and corpus processing which uses POS tagger and syntactic parser, named entity recognition using Gazetter and Transducer, ontology population, and consistency checking stages using OWL2 reasoner. 
 
 STI Innsbruck \cite{sti2013} presents the accommodation ontology. The ontology was expanded from GoodRelation vocabulary \cite{goodrelation}. It describes hotel rooms, hotels, camping sites, and other types of accommodations, their features, and
modeling compound prices as frequently found in the tourism sector. For example, the prices show the weekly cleaning fees or extra charges for electricity in vacation homes based on metered usages. Chaves et al. proposed  Hontology which is a multilingual accommodation onotology \cite{silver2012}. They divided into 15 concepts including facility, room type, train type, location, room price, etc. These concepts are similar to QALL-ME \cite{ou2009} as shown in Table \ref{tab:table1}.
Among all these, the typical properties are such as name, location, type of accommodation, facility etc. In the paper,  we use location and nearby as examples for information extraction.

\begin{table}[htbp]
    \centering
    \caption{Concept  between Hontology and QALL-ME \cite{ou2009}.}
    \label{tab:table1}
    \includegraphics[width=0.5\linewidth]{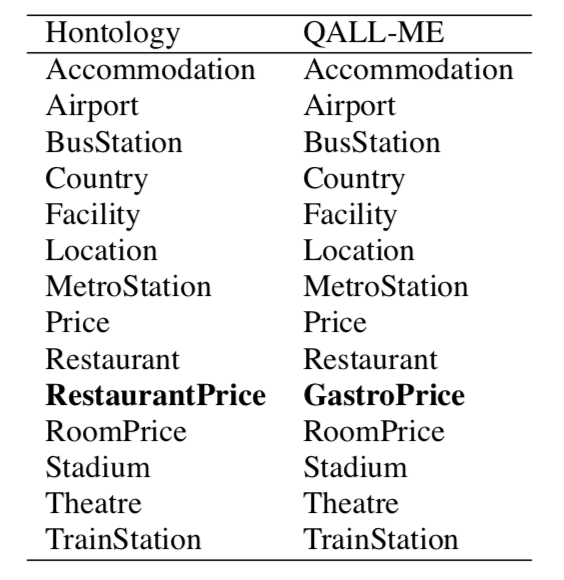}
\end{table}

\subsection{Machine Learning in NLP}
In the past, a rule-based approach is commonly used for NLP tasks such as  POS (part-of-speech), NER (named entity recognition), SBD (sentence boundary disambiguation), word sense disambiguation, word segmentation, entity relationship identification, text summarization, text classification, etc. The rule-based approach is very fragile and sensitive to individuals.
There are attempts to use
machine learning to applied to NLP tasks \cite{khan2016}.  Machine learning is used to learn language features and build a model to solve these tasks. Common models are Naive Bayes, SVM, Random Forest, Decision Tree, Markov model, etc.

In the deep learning, the use of deep network is for the purposed of learning feature.  The common model used for this task is Recurrent  Neural Network (RNN) \cite{otter2018}. The RNN captures the previous contexts as  states and is used to predict the output (such as next predicted words). RNN can be structured many ways such as  stack, grid, as well as bidirection to learn from left and right context. The RNN cell implemented can be LSTM (Long Short Termed Memory) or GRU (Gated Recurrent Unit) to support the choice of forgetting or remembering. 

One of the typical model  used is Seq2Seq model which applies bidirectional LSTM cells with attention scheme  \cite{shi2018}. The example application is abstractive text summarization \cite{yang}. On the other hand, the extractive text summarization does not employ machine learning at all. It is based on text ranking approach (similar to page ranking) to select top rank sentences \cite{Cheung2008}. The text summarization is found to be a popular application for  NLP where current approaches is still far from usable results.

Traditional Seq2Seq models have a major drawback since the pretrained weight cannot be reapplied. Recently, Google proposed  a pretrained transformer, BERT, which is bidirectional and can be applied to NLP tasks \cite{BERT}. The concept of the transformer does not rely on shifting the input to the left and to the right. The model construction also considers sentence boundaries and relationship between sentences.

NER is also another basic task that is  found to be useful for many applications such as text summarization and word relationship extraction.  To discover the named entity of a given type, lots of training data is used. The training data must be annotated with proper tags. POS tagger is the first required one since part-of-speech is useful for discovering types of entity. Other standards of tagging are IO, BIO, BMEWO, and BMEWO+ \cite{Dai2018} used  to tag positions of token inside word chunks.  Tagging process is a  tedious task where the automatic process is needed. Machine learning is therefore can be used to 
do the named entity recognition  to help automatic tagging and NER\cite{Yadav2018}

In the following,   tag labels for the named entity type we look for are defined  and we training the model for  recognizing them. Also, we can train our parser to look for the relation between our entity types. Then, we can extract the desired relation along with the named entity.

%...Typical steps for NLP processing.
%...Previous work in NLP in deep learning....LSTM-based, RNN (text generation),  

 \begin{figure}[htbp]
    \centering
    \includegraphics[width=\linewidth]{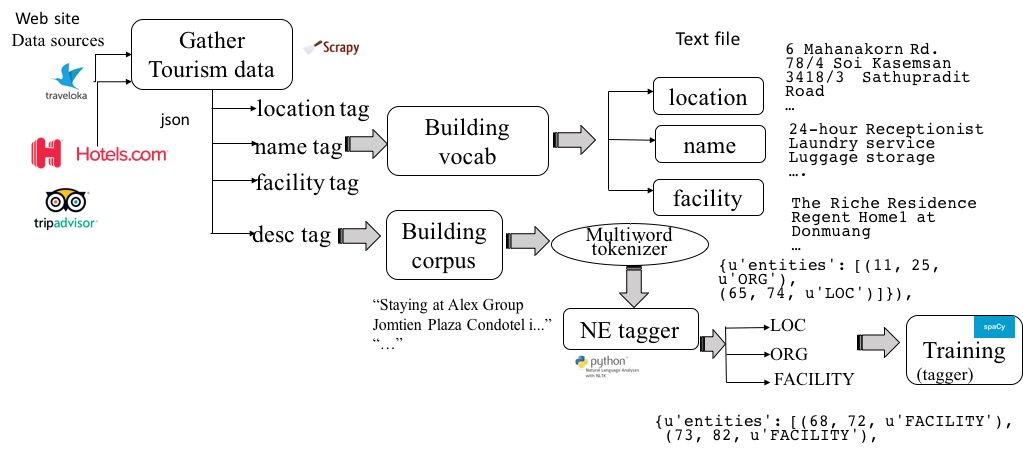}
    \caption{Overall process for creating training model.}
    \label{fig:process}
\end{figure}

%\begin{figure}[htbp]
%    \centering
%    \includegraphics[width=\linewidth]{traveloka.png}
%    \caption{Traveloka data example.}
%    \label{fig:traveloka}
%\end{figure}

\section{Methodology}

Figure \ref{fig:process} depicts the overall of the methodology for training model creation. The difficult part is the labeling the data.
 
%The labeled data is divided into 70\% for training and 30\% for testing.
%The training data is trained with spaCy (\url{https://spacy.io/usage/spacy-101}).
%The trained model is saved and reuse for testing phase.
%

%word2vec word embedding\cite{Mikolov2013}
%glove\cite{glove2014} spaCy \cite{choi2015}

\subsection{Data gathering}

To prepare the data, we  crawl data using  Scrapy library ({\small \url{https://scrapy.org/}}) in Python  going through three websites: Tripadvisor, for totally 10,202 hotels, Traveloka, for totally 5,430 hotels, and Hotels.com, for  11,155 hotels. For each website,   eight provinces are considered, including Bangkok, Phuket, Chaingmai, Phang-nga, Chonburi, Suratani, Krabi, Prachuap Khiri Khan and for each province, we collect six features, including name, description, address, facility, nearby, review.
% Which crawling into CSV file by the following in Figure \ref{fig:traveloka}

Example of the data obtained is as a json:
{\small   
\begin{verbatim} 
  [{ "address": "1/1 Moo 5 Baan Koom, Doi 
   Angkhang, Tambon Mae Ngon, Amphur Fang ,
    Mon Pin, Fang  District, Chiang Mai, 
    Thailand, 50320",
   "description": ",Staying at Angkhang Nature  
   Resort is a good choice when you are 
   visiting Mon Pin.This hotel is  ....
   "facility": "WiFi in public area,
   Coffee shop,
   Restaurant,Breakfast,...
   "name": "B.M.P. By Freedom Sky",
   "nearby": "Maetaeng Elephant Park,
   Maetamann Elephant Camp,Mae Ngad Dam 
   and Reservoir,Moncham",
  "review": "" }
  { ...}]   
\end{verbatim}}

\subsection{Corpus building}

Next, we build the vocabularies which  are special keywords in our domain, saved in text files. These keywords can be multiple words or chunks. They are used to tokenize word chunks (multiword tokenizer) for a given sentence. 
For example, Thong Lor, JW Marriott, Room service etc.
they each should be recognized as a single chunk. To build our multiword vocabulary files we extract the values from the json field: `name', `location', `nearby'  and `facility'.

\begin{table}[htbp]\small
    \centering
        \setlength\tabcolsep{2pt}
    \caption{Corpus statistics.}
    \label{tab:stats}
    \begin{tabular}{|c|c|c|c|c|c|}\hline
         location&facility&nearby&hotel name& all & all(nodup) \\\hline
         17,660 & 17918& 18,822 & 41,168&58,828 &37,107\\\hline 
    \end{tabular}
    
\end{table}
 
  The values in these fields, separated by comma are split and saved into each text file: 
1) location, and nearby are saved as  {\small \texttt{location-list.txt}}
2)  name is saved as {\small \texttt{hotel-name-list.txt}}
3) facility is saved as {\small \texttt{facility-list.txt}}, as in Figure \ref{fig:process} in "Building vocab".
The number of values for keywords for each json field and total vocabularies for each field are displayed in Table \ref{tab:stats}.  Column `all (nodup)' shows the the all combined vocabularies when removing duplicates.
Then, multi-word expression tokenizer is built using Python library with  \texttt{nltk.tokenize.mwe}.

\subsection{Data preparation and model creation}
For creating training data for spaCy ({\small \url{http://spacy.io}}), we have to build the training data  in a compatible  form as its input. Depending on the goal of training model, the label input data are formed properly.  The interested named entities are LOC, ORG, and  FACILITY. LOC refers to location or place names. ORG refers to hotel or accommodation names. FACILITY refers to facility types.  LOC and ORG are the built-in entity where spaCy is already trained. However, we have to add our vocabulary keywords in since our location name and hotel name (considered as ORG) are specific to our country.
For FACILITY, we create our new named entity label for the name of facility in the hotel.

The model is built for recognizing these new entities and keywords.
 It is trained  with our location name, hotel name and facility. 
 To  build the training data from our json files,
  we extract sentences  manually  from description fields. We cannot use the whole paragraph and annotate  them since each paragraph is too long and there are lots of irrelevant sentences/words which can create a lot of noises in training data.
Thus, we have to specifically find the sentences that contain these named entities.
The sentences are then selected  manually from the paragraphs which contains either hotel names, locations, facilities.

The selected sentences
are tokenized the sentences and  the named entities, according to the corpus are searched for and tagged as LOC, ORG, or FACILITY. The   tuple (starting index position, ending index position) for each word chunk in each  sentence is recorded for each tagged chunk. 

Example of tagging is as following:
 {\small
\begin{verbatim}
'text': Staying at @ Home Executive Apartment
 is a good choice when
  you are visiting
   Central Pattaya.
'entities': [(11,37,ORG), (77,92,LOC)]
\end{verbatim}}
 
In this sentence, the text is the extracted sentence from the paragraph.   There are two entities, `@ Oasis Resort' which begins at character index 11 ends at index 37 and its type is ORG name (hotel name). `Wang Pong' begins at character index 77 ends at index 92 and its type is LOC.

\begin{figure}
    \centering
    \includegraphics[ height=1.7in]{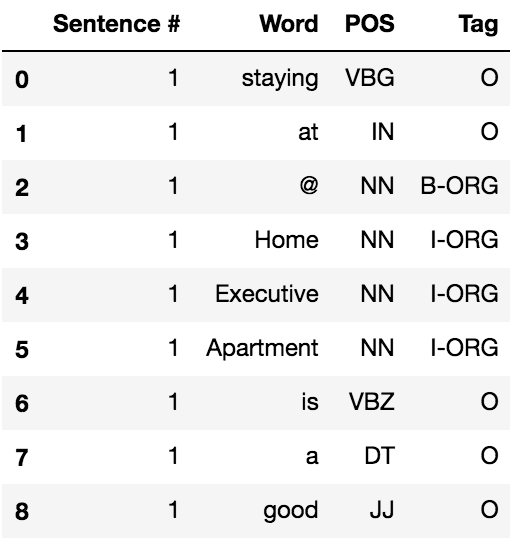}
    \caption{BERT training data for NER}
    \label{fig:bert1}
\end{figure}

For BERT, the training data is adjusted as in   Figure \ref{fig:bert1}. 
We transform the input raw corpus from spaCy to its form.  
The sentence  is split into words and each word is marked with POS tagger. Column `Tag' is our tag name for each word. We have to split the multiword named entity and tagging each word since we have to use BERT tokenizer to convert to word ID before training.
In the figure, our hotel name starts at `@', we use the tag B-ORG (begining of ORG)
and `Home' is the intermediate word after it (I-ORG).  BERT needs to convert the input sentences into  lists of word IDs along with the list of  label IDs.
The lists are  padded  to equal size before sending  them to  train.  

\subsection{Train new word embedding}
Since we have our new vocabularies  for location name, hotel name, and facility, creating representation for these vocabularies will be useful for other related NLP tasks. Available word representations are such as word2vec in  {\small\url{https://pypi.org/project/gensim/}}, GloVe, containing around 400,000 vocabularies \cite{glove2014}, ConceptNet -NumberBatch\cite{conceptnet} (containing around 484,556 vocabularies), GoogleNews  model \cite{googlenews} (containing 3,000,000 vocabularies) etc.  

The existing representation vector for each word can be of varied length: for GloVe 50,100,200,300 , and for BERT, 768 etc. 
The approach to generate embedding for the new word chunks are 
1)  tokenize the new raw corpus  using our multiword vocabularies
2) create a set of vocabularies (\emph{vocab}) from  the new raw  corpus, (the word is in a vocabulary set if it has a number of occurrences in the document greater than the threshold.) 
3) adopt pretrained word model, from existing representation e.g., GloVe, or GoogleNews, etc.
4) intersect \emph{vocab}  and the existing vocabularies  in pretrained word model to adopt pretrained weight,
5) train the word model for the \emph{vocab}.

In our case, we train the embedding from the field `desc' data (called raw corpus) of the json file for our new keywords. Totally, raw corpus contains 5,660,796 characters, 873,682 tokens. The raw corpus is cleaned by removing punctuations and made as lower case just like the multiword vocabularies. Each sentence from raw corpus is tokenized and the number of vocabularies obtained  is 2,818 (with the minimum frequency of 7). 
Finally, we have the new set of vocabularies   including our new names, locations, facilities.

\section{Experiments}
 
The experiments are run on Intel 2.6 GHz  Core i5
RAM 8 GB for training the models.
We report the results into three subsections: The accuracy of NE recognition for spaCy, the accuracy of BERT recognition and the similarity scores of your vocab corpus, then we explain additional pipelines.

All the code and data can be downloaded at  {\small\url{http://github.com/cchantra/nlp_tourism}.
\subsection{SpaCy NE and BERT}
We compare three models in building using  spaCy.
1) the  model  to recognize ORG/LOC, 2)  the model recognize FA CITY and 3) the model to recognized all  ORG/LOC/FACILITY.
For model 1), we use raw corpus with annotations only for ORG/LOC, containing 13,019 sentences.
For 2),  the raw corpus only contains FACILITY sentences, totally, 13,522 rows. For 3), we combine the corpus from 1)+2) to create one model recognizing all entities.

Figure \ref{fig:loss} presents loss of the training phase for three models of spaCy.
\begin{figure}[htbp]
    \centering
    \includegraphics[ height=2in]{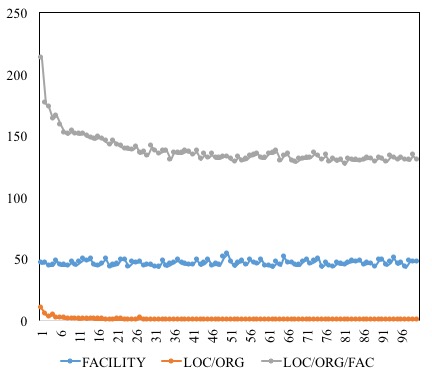}
    \caption{Loss values for three models}
    \label{fig:loss}
\end{figure}

Training for LOC/ORG has no difficulty at all since ORG/LOG labels are already in the pretrained model of spaCy. We  add new words with this label. For FACILITY, we add it as the new tag label and   the model  is trained to learn this new tag.
The loss is higher than LOC/ORG model. Also, when we combine three tags, the total loss are even higher. Note that if we do not manually select sentences containing these  named entities, the large raw corpus takes long time to train and the loss is over 1,000.

\begin{table}[htbp]

    \centering\small
     \caption{Comparison between predicted and annotated NE using spaCy.}
    \label{tab:compare1}
    \setlength\tabcolsep{2pt}
    \begin{tabular}{|c|c|c|c|c|c|c|} \hline
        Type&\multicolumn{2}{c|}{LOC/ORG}&\multicolumn{2}{c|}{FAC}&\multicolumn{2}{c|}{LOC/ORG/FAC}\\\cline{2-7}
        & train&test&train&test&train&test\\\hline
         \#Annotated&	46,327&	19,787&	22,167&	9,427&	93,661&	40,745\\\hline
         \#Predicted	&70,156	&29,873&	18,716&	7,299&	85,867&	37387\\\hline
Diff&	23,829&	10,086&	-3,451&	-2,128&	-7,794&	-3,358\\\hline
Accuracy (\%) &151.43 &	150.97 &	 84.43 &	77.42& 91.67 &	91.75\\\hline
    \end{tabular}

\end{table}
The  tagged data is divided into 70\% training and 30\% testing. 
Next, Table \ref{tab:compare1} shows the correctness between three models. Row `Annotated' is our manual label counts and `Predicted' is the number of NEs predicted. For LOC/ORG, spaCy has the built-in labels for it; thus, it discovers more entities than annotated labels. Thus, the accuracy is higher than 100\%. 
For `FACILITY', the number of predicted labels missed is  around 15\% for training and 22\% for testing. The accuracy is 84\% and 77\% for training and testing respectively. For LOC/ORG/FAC, the  number of missing ones are 8\% for training and 8\% for testing. The accuracy is around 91 \% for both training and testing.

%Figure \ref{fig:spacy} shows the graph comparison between predicted named entities and annotated named entities for each case.
% 
%\begin{figure}[htbp]
%    \centering
%    \includegraphics[width=3in]{graph.jpg} 
%    \caption{spaCy NE models' correctness}
%    \label{fig:spacy}
%\end{figure}
 
 For BERT, the training accuracy is depicted in  Table \ref{tab:compare2-ownword}, BERT   performs well  on all the tasks.
 Origina BERT relies on its tokenizer which includes the token conversion to ID.
  It cannot tokenizer our special multiword name which are proper noun very well.  
  For example, `Central Pattaya' is tokenized into
 `u'central', u'pat', u'\#\#ta', u'\#\#ya'.
 which makes our labels wrong starting at  position `pattaya'.
  The unknown proper nouns are chopped into portions which makes the labels shift out. According to the paper's suggestion, this needs to be solved by using own wordpiece tokenizer \cite{wordpiece2016}. 
  %For facility name, it almost has no problem with the proper noun phase, yielding  the highest accuracy.
 We then create our set of own words for training based on raw corpus. The accuracy results are measured based on the number of correct predictions.
 The results for all cases are around 70\%
 
% \begin{table}[htbp]
%
%    \centering\small
%     \caption{Comparison between predicted and annotated NE using BERT.}
%    \label{tab:compare2}\small
%    \setlength\tabcolsep{2pt}
%    \begin{tabular}{|l|c|c|c|c|c|c|} \hline
%        Type&\multicolumn{2}{c|}{LOC/ORG}&\multicolumn{2}{c|}{FAC}&\multicolumn{2}{c|}{LOC/ORG/FAC}\\\cline{2-7}
%        & train&test&train&test&train&test\\\hline
%  Loss	&0.061	&0.153&0.03	&0.0295	&0.123	&0.109 \\\hline
%Accuracy&	0.958	&0.955	&0.939	&0.93	&0.866	&0.866 \\\hline
%F1	&0.737	0.715&	&0.419	&0.43	&0.391	&0.464\\\hline
%    \end{tabular}
%    \end{table}

 \begin{table}[htbp]

    \centering\small
     \caption{Comparison between predicted and annotated NE using BERT.}
    \label{tab:compare2-ownword}\small
    \setlength\tabcolsep{2pt}
    \begin{tabular}{|l|c|c|c|c|c|c|} \hline
        Type&\multicolumn{2}{c|}{LOC/ORG}&\multicolumn{2}{c|}{FAC}&\multicolumn{2}{c|}{LOC/ORG/FAC}\\\cline{2-7}
        & train&test&train&test&train&test\\\hline
  Loss	& 0.019	&0.0165&	0.375	&0.029	&0.123	&0.109 \\\hline
Accuracy&	0.751	&0.717	&0.859 &	0.7484& 	0.736&	0.629 \\\hline
F1	& 0.258	&0.346	&0.245&	0.245	&0.287	&0.464\\\hline
    \end{tabular}
    \end{table}
% 
% Figure \ref{fig:loss_bert}. The accuracy is measured by the number of the correct predictions and labels. The accuracy is high
% \begin{figure}[htbp]
%    \centering
%    \includegraphics[width=3in]{loss_bert.jpg} 
%    \caption{BERT NE model correctness}
%    \label{fig:loss_bert}
%\end{figure}

\subsection{Extracting training sentences}
When crawling the data source, we intend to obtain a list of paragraph describing a particular hotel or hotel review. These information are useful for creating corpus and review summary.

Text summarization is a common
techniques used in on basic NLP tasks as well. The extracted summary can be either from abstractive or extractive method. For our task to extract sentences for creating training raw corpus for named entity training. We can use extractive summary to pull important sentences and use these sentences as our raw corpus. The pipeline for raw corpus building is modified in Figure \ref{fig:textrank}.  In this figure, TextRank is used to pull the important sentences.

\begin{figure}[htbp]
    \centering
    \includegraphics[width=  \linewidth]{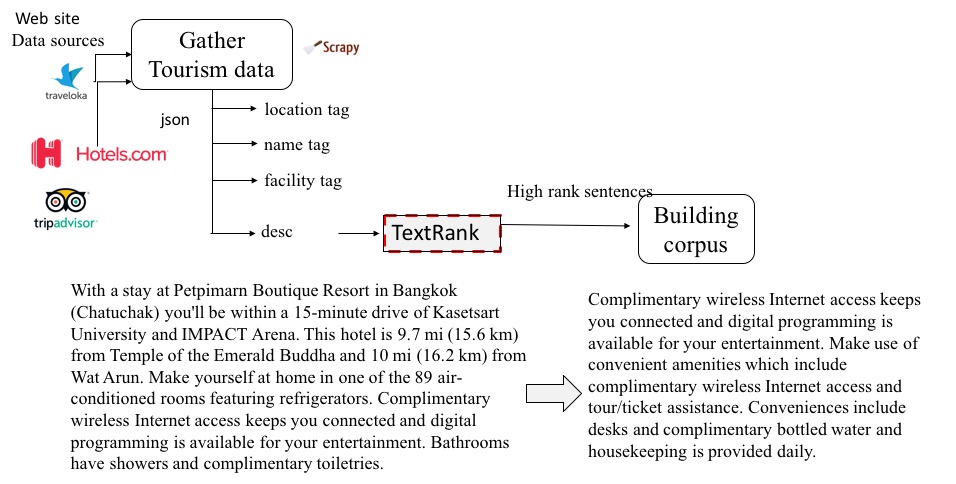}
    \caption{Adding pipeline of TextRank.}
    \label{fig:textrank}
\end{figure}

\subsection{Noun chunk addition}
We can use spaCy noun chunk to split sentences
into noun phrases. The noun phrases can be selected 
and added to keyword textfile. Then they are used together with multiword tokenizer. The additional step is depicted in 
Figure \ref{fig:nounchunk}.

\begin{figure}[htbp]
    \centering
    \includegraphics[width=\linewidth,height=1.5in]{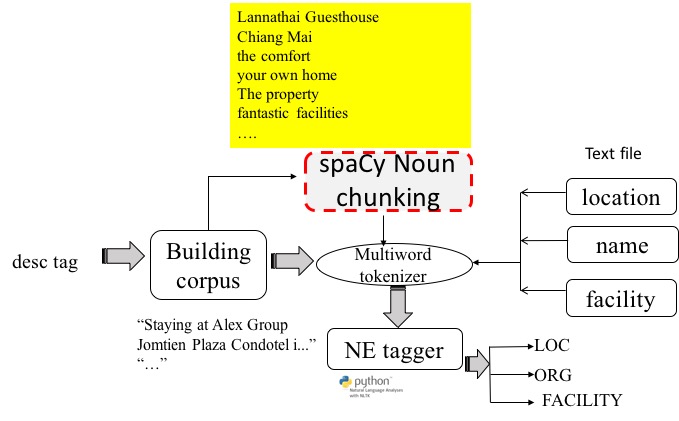}
    \caption{Adding pipeline of nounphase chunking.}
    \label{fig:nounchunk}
\end{figure}

\subsection{Relation type extraction}
After obtaining named entities, we also would like to find out the relationship between named entities in a sentence. We can train the model to recognize the relationship between two words. In training relation using spaCy,  the relationship between the words  is defined as dependency (deps)  as in the following.
{\small
\begin{verbatim}
("Conveniences include desks and complimentary
bottled water", 
{'heads': [0, 0, 0, 0, 6, 6,0], 
# index of token head and
'deps': ['ROOT', '-', 'FACILITY', 
'-', 'TYPE', 'TYPE','FACILITY']   
# dependency type between pair
 }),
….
\end{verbatim}}

In this example, `heads'  is a list whose  length equals to number of words.  `deps' is the list of name relations. Each element refers to its parents. For example, 0 at the last element (water)  refers to the relation `FACILITY' to `convenience' and 6 refers that it is the modifier `TYPE' of `water'.  `-' means no relation (or ignored).  Obviously, it requires efforts for manual annotations. 
Figure \ref{fig:relation1}
 shows the example of  relation in our raw corpus. The relation exhibits the property {\small\texttt{isLocated} }used in tourism ontology concept.
 
 In Figure \ref{fig:relation2}, there are lots of relations in one sentences, making it harder to annotate.  The sentence exhibits  both {\small\texttt{isLocated}} and {\small \texttt{hasFacility}} properties.
 Figure \ref{fig:relation_create} is the new pipeline following NE recognition stage to train relation model.
We extract only sentences with found NEs and start annotate relation dependency for each sentence.

\begin{figure}[htbp]
    \centering
    \includegraphics[width=\linewidth]{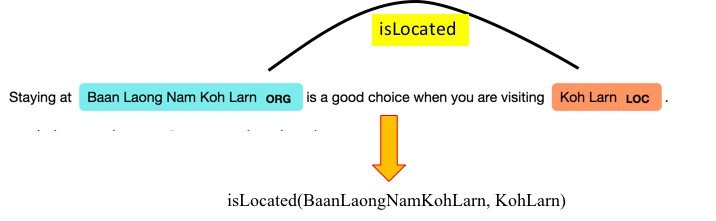}
    \caption{Relation capturing example 1}
    \label{fig:relation1}
\end{figure}

\begin{figure}[htbp]
    \centering
    \includegraphics[width=\linewidth]{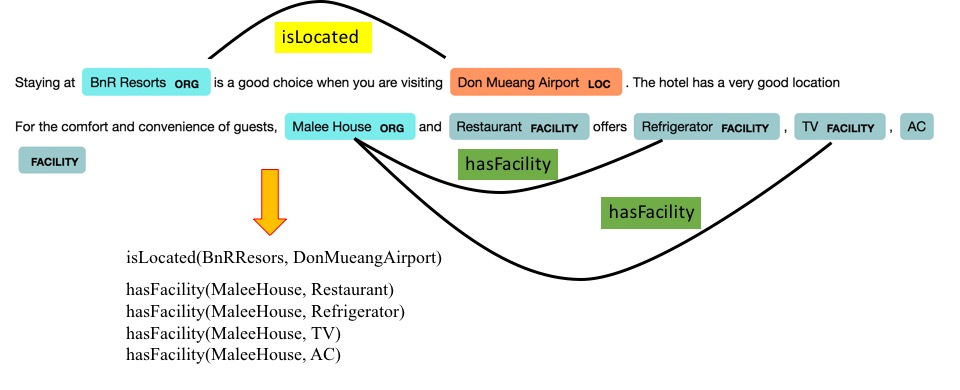}
    \caption{Relation capturing example 2}
    \label{fig:relation2}
\end{figure}

\begin{figure}[htbp]
    \centering
    \includegraphics[width=\linewidth]{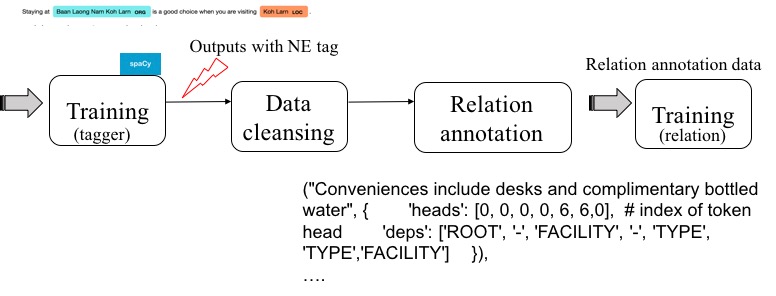}
    \caption{Relation creation model}
    \label{fig:relation_create}
\end{figure}

\begin{table}[t]
    \centering
    \setlength\tabcolsep{.0pt}
    \setlength\extrarowheight{-9pt}
    \caption{Comparison of similarity scores.}
    \label{tab:word2vec}
     \small
   
    \begin{tabular}{|c|p{1.3cm}|p{1.5cm}|p{1.3cm}|p{1.5cm}|p{1.4cm}|}\hline
        \multicolumn{6}{|c|}{\cellcolor{blue!25}NEAR}\\\hline
        vocab &('airport', 0.717 )&
 ('don mueang international airport dmk', 0.662)&
 ('only', 0.640)&
 ('bs', 0.604)&
 ('mo chit bts station', 0.587)\\
 \hline
      GloVe-50& \cellcolor{yellow!25}('nearby', 0.924)&
 ('town', 0.880)&
 ('area', 0.877)&
 ('vicinity', 0.818 )&
 ('located', 0.818 ) \\\hline

        GloVe-300&  \cellcolor{yellow!25}
        ('nearby', 0.751)&
 ('located', 0.637)&
 ('town', 0.632)&
 ('vicinity', 0.627)&
 ('area', 0.576)\\\hline
        ConceptNet&('motel', 0.464)&
 ('deal', 0.331)&
 ('sai', 0.322)&
 ('positioned', 0.310)&
 ('kitchens', 0.299)\\\hline
        GoogleNews &  \cellcolor{yellow!25}
        ('nearby', 0.606)&
 ('vicinity', 0.559)&
 ('west', 0.525)&
 ('at', 0.517)&
 ('located', 0.457)\\\hline
      \multicolumn{6}{|c|}{\cellcolor{blue!25}Bangkok}\\\hline
      vocab &
      ('louis tavern hotel', 0.782)& ('donmuang', 0.719),&('suite', 0.715)& ('don muang', 0.705)& ('regent', 0.691)\\\hline
      GloVe-50 &
      ('malaysia', 0.711)&  \cellcolor{yellow!25} ('thai', 0.688 )& ('seoul', 0.683 )& ('airport', 0.642)& ('pattaya', 0.639)\\\hline
      GloVe-300&   \cellcolor{yellow!25}
      ('thai', 0.602)& ('seoul', 0.507)& ('phuket', 0.483)& ('pattaya', 0.465)& ('suvarnabhumi', 0.408 )\\\hline
      ConceptNet &
      ('staffed', 0.426)& ('tub', 0.421)& ('sovereign', 0.401)& ('additional', 0.396)& ('rainfall', 0.386) \\\hline
      GoogleNews &  \cellcolor{yellow!25}
      ('thai', 0.575)& ('argentina', 0.485)& ('pai', 0.477)& ('malaysia', 0.468 )& ('anna', 0.462) 
      \\\hline
    \end{tabular}
    
\end{table}

\subsection{Adding similar word}
To extract relations, the keywords indicated specific relations must be gathered. One can define a set of words and use word similarity to help find out other words to add to the relation keywords as
in Figure \ref{fig:word2vec}.

Table \ref{tab:word2vec} displays effectiveness of different representations: GloVe, ConceptNet-NumberBatch, GoogleNews.
We add our new vocabularies to these existing vocabularies and compared the similarity score for each given word.
In row 'vocab', it is the representation trained by only our crawled raw corpus.  Row 'GloVe-50', we use  the representation from GloVe 50 ({\small\texttt{glove.6B.50d.txt}}) to perform the training for additional vocabularies.
Similarly, for 'GloVe-300' ({\small\texttt{glove.6B.300d.txt}}), 'ConceptNet', and 'GoogleNews', we adopt pretained representation to train new vocabularies from the raw corpus. We can see that the most similar word to 'Near' is 'nearby' using our raw corpus  alone will not give the right representation. GloVe and GoogleNews give
  the better similarity score.
  In the second example, for the word 'Bangkok', GloVe-300 and GoogleNews give 'thai' which seems to me a good similar word.
  
   \begin{figure}[htbp]
    \centering
    \includegraphics[width=\linewidth]{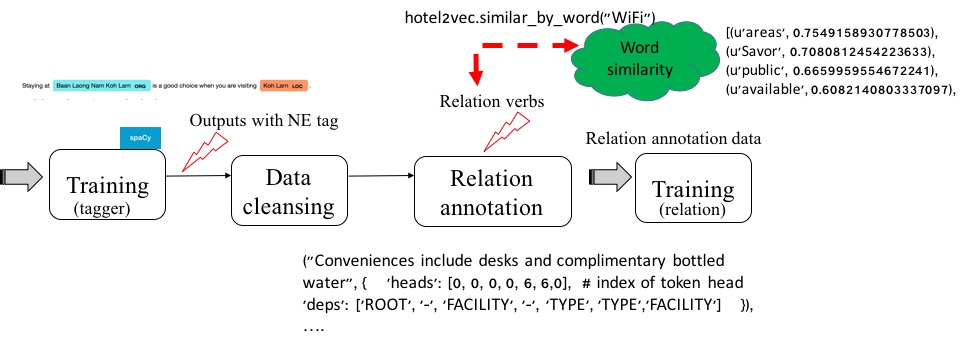}
    \caption{Adding word similarity pipeline}
    \label{fig:word2vec}
\end{figure}
  \section{Conclusion and Future Work}
This paper presents a methodology for extracting from unstructure information   focusing on tourism data.  We demonstrate the prototype of    location and facility extraction of hotel accommodations using machine learning.  The major demonstration is based 
 the preprocessing part which builds the specific vocabularies and raw corpus  and annotations, necessary for
  the model  construction. The model is to learn  to recognize named entities and relations.   Two approaches for building models, spaCy and BERT are discussed. Also,   the construction of new  tourism vocabulary word2vec representation  and its use for other NLP tasks in the domain are also presented.

The methodology can be generalized to  recognize entities of other domains. Several other machine learning models can be handy tools for preprocessing training data.
For the tourist review information,   the extractive
approach can be used to extract the highlight of the user review and abstractive approach can be used to build the review summary. The abstractive approach replies on word2vec representation (with new vocabularies). 
Also, we can build the model to identify the positive and negative review.
Document classification can be used to classify the type information containing in each sentence before sending it to build the raw corpus the named entity extraction.

\addtolength{\textheight}{-12cm}   % This command serves to balance the column lengths

%%%%%%%%%%%%%%%%%%%%%%%%%%%%%%%%%%%%%%%%%%%%%%%%%%%%%%%%%%%%%%%%%%%%%%%%%%%%%%%%

%%%%%%%%%%%%%%%%%%%%%%%%%%%%%%%%%%%%%%%%%%%%%%%%%%%%%%%%%%%%%%%%%%%%%%%%%%%%%%%%

%%%%%%%%%%%%%%%%%%%%%%%%%%%%%%%%%%%%%%%%%%%%%%%%%%%%%%%%%%%%%%%%%%%%%%%%%%%%%%%%

\small
\section*{ACKNOWLEDGMENT}

The work is supported in part by Kasetsart University Research and Development Institute, Bangkok, Thailand.

%%%%%%%%%%%%%%%%%%%%%%%%%%%%%%%%%%%%%%%%%%%%%%%%%%%%%%%%%%%%%%%%%%%%%%%%%%%%%%%%
 \small
\bibliographystyle{IEEEtran}
\bibliography{IEEEabrev,ref}

\end{document}